# Longitudinal Missing Data Imputation for Predicting Disability Stage of Patients with Multiple Sclerosis

Mahin Vazifehdan[*,1], Pietro Bosoni[1], Daniele Pala[1], Eleonora Tavazzi[2], Roberto Bergamaschi[2], Riccardo Bellazzi[2], Arianna Dagliati[2]

[1] Department of Electrical, Computer and Biomedical Engineering, University of Pavia, Pavia, Italy.
[mahin.vazifehdan01; pietro.bosoni02]@universitadipavia.it, [daniele.pala; riccardo.bellazzi; arianna.dagliati]@unipv.it

[2] IRCCS Mondino Foundation, Pavia, Italy. [eleonora.tavazzi; roberto.bergamaschi]@mondino.it

[*]corresponding author

*Keywords: Longitudinal Missing Data, Multiple Sclerosis, Multiple imputations, Joint Modelling Imputation*

**Abstract.** Multiple Sclerosis (MS) is a chronic disease characterized by progressive or alternate impairment of neurological functions (motor, sensory, visual, and cognitive). Predicting disease progression with a probabilistic and time-dependent approach might help in suggesting interventions that can delay the progression of the disease. However, extracting informative knowledge from irregularly collected longitudinal data is difficult, and missing data pose significant challenges. MS progression is measured through the Expanded Disability Status Scale (EDSS), which quantifies and monitors disability in MS over time. EDSS assesses impairment in eight functional systems (FS). Frequently, only the EDSS score assigned by clinicians is reported, while FS sub-scores are missing. Imputing these scores might be useful, especially to stratify patients according to their phenotype assessed over the disease progression. This study aimed at i) exploring different methodologies for imputing missing FS sub-scores, and ii) predicting the EDSS score using complete clinical data. Results show that Exponential Weighted Moving Average achieved the lowest error rate in the missing data imputation task; furthermore, the combination of Classification and Regression Trees for the imputation and SVM for the prediction task obtained the best accuracy.

## 1 Introduction

Multiple Sclerosis (MS) is a chronic and progressive neurological disease affecting the central nervous system due to an immune system error that damages the protective covering of nerve fibres, called myelin sheath, leading to various symptoms such as cognitive impairment, movement and coordination difficulties [1, 2, 3]. The growing trend of exploiting longitudinal clinical data aims at improving diagnostic accuracy and tailoring treatment plans [4]. Longitudinal studies take repeated measures of a set of variables from the same group of subjects over time at several points on a time grid, which might be irregular. In addition, a significant challenge is posed by missing data; omitting incomplete observations may lead to the loss of a considerable amount of data, which can be particularly problematic in medical applications where data collection is time-consuming [5]. Moreover, most machine-learning approaches are susceptible to incomplete datasets. Therefore, imputing missing data is one of the preferred approaches for dealing with missing data, which involves predicting missing values based on the available data, resulting in a complete dataset for further analyses [6].

The Expanded Disability Status Scale (EDSS) is a clinical score that describes the severity of MS patients' disability on a scale from 0 to 10 in 0.5-unit increments, with higher scores indicating greater levels of neurologic impairment [7]. The EDSS score is based on the assessment of eight functional systems (FS), such as pyramidal (muscle weakness or difficulty moving limbs), cerebellar (ataxia, loss of balance, coordination, or tremor), brain stem (problems with speech, swallowing, and nystagmus), sensory (numbness or loss of sensations), bowel and bladder, visual (problems with sight), cerebral (problems with thinking and memory), and other functions. Each FS is scored on a scale of 0 (no disability) to 5 or 6 (more severe disability). Often, only the EDSS score assigned by clinicians is reported, while FS sub-scores are missing.

This work aims at i) imputing missing FS sub-scores by comparing state-of-the-art imputation methods for longitudinal data, including demographic and FS sub-scores features; ii) predicting the EDSS disability stage using complete longitudinal data. It is important to note that while the multiple imputation framework typically involves training a statistical model on multiple imputed datasets and then pooling the estimates [8], our main goal is to obtain a reliable and complete dataset that can be used for further analyses, including unsupervised approaches for patient stratification. In addition, another aspect of our study is to evaluate Joint Modelling (JM) with clustered and single levels data. For the first task, we compared the error rates of different imputation methods in a classifier-independent scenario. For the second task, we evaluated the predictive accuracy of well-known prediction methods, specifically k-Nearest Neighbor (KNN), Light Gradient Boosting (LightGBM), Random Forest (RF), and Support Vector Machine (SVM) on a complete dataset.

## 2 Data and Methods

### 2.1 Dataset

We analysed longitudinal clinical data of MS patients treated at the MS Center of the Mondino Foundation (Pavia, Italy) and collected via the Italian MS register. The dataset includes 14,226 follow-up records of 919 patients with 12 features including demographic and FS sub-scores. The EDSS score evaluated by the clinicians is assumed as the target class. Table 1 illustrates the missing rate of each feature.

*Table 1: Missing rate of dataset features*

| Feature Group | Feature Name | Missing rate (%) |
|---|---|---|
| Static features | Sex | 0 |
| | Age | 0 |
| | MS in pediatric age | 0 |
| FS sub-scores | Pyramidal | 11.28 |
| | Cerebellar | 13.27 |
| | Tronchioencephalic | 13.22 |
| | Sensitive | 12.77 |
| | Sphincteric | 13.53 |
| | Visual | 13.70 |
| | Mental | 13.84 |
| | Deambulation | 14.73 |
| Target feature | EDSS score | 0 |

*2.2 Imputation Methods*

A total of 14 imputation methods suitable for longitudinal data were employed to complete the dataset, and statistical analysis was performed to identify the top five imputation methods. These imputation methods can be categorized into single and multiple imputations methods. Regarding single imputation, we employed Linear and Spline interpolation, Last Observation Carried Forward (LOCF), and Exponential Weighted Moving Average (EWMA). For multiple imputations, we used Predictive Mean Matching (PMM), Classification and Regression Tree (CART), Bayesian Linear Regression (BLG), Random Forest, different versions of Linear Regression Model, and JM of Single- and Clustered-Level data. JM is a statistical method used to impute missing data in both single and multi-level/hierarchical data structures, and it is particularly useful in clinical longitudinal studies where repeated measurements are taken on individuals [9]. This method assumes that the incomplete variables follow a multivariate normal distribution and uses a joint multivariate model for partially observed data. It generates imputations using a Gibbs sampler, updating the covariance matrix with a Metropolis-Hastings step. All imputation methods were performed in R version of 4.2.3 (using Jomo [9], Mice [10], and imputeTS [11] packages).

To evaluate the imputation methods' performances, a complete dataset was assembled by considering only the instances without missing data. The complete dataset was then altered by randomly removing records of attributes. Removed values were imputed with the selected methods, which were assessed by the Root Mean Squared Error (RMSE) calculated as in Equation 1:

$$RMSE = \sqrt{\frac{\sum_{i=1}^{N}(x_i - \hat{x}_i)^2}{N}} \quad (1)$$

where $x_i$ is the actual value, $\hat{x}_i$ is the imputed value, and $N$ is the total number of missing values observed in the dataset.

*2.3 Predictive Models*

For the prediction task, we considered four individual and ensemble learning methods, specifically KNN [12], LightGBM [13], RF [14] and SVM [15] which are representative methods for neighbour-based, ensemble learning and kernel-based methods. For evaluating the prediction methods, we used the coefficient of determination ($R^2$), computed as in Equation 2:

$$R^2 = 1 - \frac{SSR}{SSt} \quad (2)$$

where $SSR$ is the sum of the residuals squared and $SSt$ is the sum of the squared distance of the data from the mean.

We divided the original incomplete longitudinal dataset into training (80%) and test (20%) sets, considering the patients' identifier in the splitting procedure: observations belonging to a specific patient were assigned to either the train or test set, ensuring that no observations from the same patient were distributed across both sets. We employed a 10-fold Cross Validation (CV) on the incomplete training data to determine both the best combination of imputation and predictive models and the models hyperparameters using a grid search. This procedure aimed to identify the optimal imputation-prediction pairs from all the possible combinations within each fold. For each pair, we calculated the mean $R^2$ value over each CV iteration. The best five pairs of imputation and predictive models, along with their respective optimal hyperparameters, were selected to first impute the missing test data, and then predict the final outcomes on the test set.

## 3 Results

### 3.1 Imputation Methods Performance

Table 2 displays the performances of the implemented imputation methods. Multiple imputations and joint modelling approaches were initially expected to yield superior outcomes, but results indicate that single imputation techniques as Linear Interpolation and EWMA allowed to obtain the lowest RMSE. The effectiveness of EWMA can be attributed to its ability to assign varying weights to individual observations over time, leading to better results than Linear, Spline, and LOCF methods. The differences among the performances of interpolation methods suggest that our longitudinal data is relatively smooth, and the variations between points are not extreme.

*Table 2: RMSE results of imputation methods (in bold the best values).*

| Imputation Group | Method | RMSE |
| --- | --- | --- |
| Single Imputation | Linear | **0.48** |
|  | Spline | 0.62 |
|  | LOCF | 0.61 |
|  | EWMA | **0.47** |
| Multiple Imputations | PMM | 0.69 |
|  | CART | **0.55** |
|  | RF | **0.57** |
|  | BLG | 0.76 |
|  | LG | 0.96 |
|  | LGP | 0.62 |
|  | LGnob | 0.75 |
| Joint Modelling | JM Clustered-Level data | **0.60** |
|  | JM Single-Level data | 0.75 |
|  | JM LG | 0.76 |

LOCF: Last Observation Carried Forward; EWMA: Exponentially Weighted Moving Average; PMM: Predictive Mean Matching; CART: Classification and regression tree; RF: Random Forest; BLG: Bayesian Linear Regression; LG: Linear Regression; LGP: Linear Regression through Prediction; LGnob: Linear Regression without Parameter Uncertainty; JM LG: Joint Modelling with Linear Regression.

Tree-based approaches, such as CART and RF, yielded the best performance within the Multiple Imputation group. Interestingly, the number of trees employed for imputing missing data did not significantly impact performance in our dataset, since the RF method employed ten trees, while the CART method utilized only one tree.

Although Multiple Imputation with various versions of Linear Regression methods resulted in minor outcomes compared to tree-based methods, Linear Regression without parameter uncertainty (LGnob) achieved acceptable results; this could indicate that the relationship between variables is linear, and there is no uncertainty in the estimated coefficients of the model equation. Regarding Joint Modelling group, our findings suggest that JM with Clustered-Level data outperformed JM with Single-Level data, mainly because the nature of longitudinal data is two-level.

*3.2 Predictive Methods Performance*

Table 3 presents the results obtained on both training and test sets. RF and SVM outperformed the other predictive models; in particular, SVM achieved the highest score with an $R^2$ value of 0.891 when using the training data imputed with the CART method. The CART-SVM obtained the best performance also in the test set with an $R^2$ value of 0.725. The RF predictive method showed good performances across three of the five imputation methods. In addition, our findings revealed that JM with Clustered-Level data imputation approach produced acceptable results when combined with the RF predictive model, achieving an $R^2$ score of 0.720 and 0.622 with the training and test sets, respectively.

Regarding the other performances on the training set, KNN gave different outcomes when used with different imputation techniques. Among these techniques, CART emerged as the most effective method for both KNN and LightGBM. Notably, the adoption of the CART imputation method consistently resulted in $R^2$ scores over 0.82 for all prediction methods, highlighting the effectiveness of combining CART with predictive approaches.

*Table 3: Prediction results with imputed datasets on training and test sets (in bold the best values)*

| Imputation Method | Prediction Method | $R^2$ score (Train set) | $R^2$ score (Test set) |
|---|---|---|---|
| Linear | KNN | 0.621 | - |
| | LightGBM | 0.715 | - |
| | RF | **0.742** | 0.688 |
| | SVM | 0.720 | - |
| EWMA | KNN | 0.616 | - |
| | LightGBM | 0.696 | - |
| | RF | **0.715** | 0.673 |
| | SVM | 0.710 | - |
| CART | KNN | 0.830 | - |
| | LightGBM | 0.881 | - |
| | RF | 0.887 | - |
| | SVM | **0.891** | **0.725** |
| RF | KNN | 0.799 | - |
| | LightGBM | 0.848 | - |
| | RF | 0.855 | - |
| | SVM | **0.856** | **0.709** |
| JM Clustered-Level data | KNN | 0.628 | - |
| | LightGBM | 0.706 | - |
| | RF | **0.720** | 0.622 |
| | SVM | 0.698 | - |

## 4 Conclusion

In this study, we addressed two primary objectives. The first aim involved analyzing the performance of 14 different imputation methods, which were categorized into three major groups suitable for longitudinal data: single, multiple, and joint modelling imputation strategies.

The second aim focused on predicting the EDSS disability stage of patients with MS using the imputed datasets. The best five imputation methods were chosen based on the RMSE score; afterwards, the imputed datasets were utilized for predicting the patient's disability stage using four prediction methods (KNN, LightGBM, RF, and SVM), considering all the possible combinations of imputation and prediction methods to select the best pairs. The EWMA method showed the most effective imputation performance, outperforming other robust imputation methods such as the JM with Clustered-Level algorithm; however, CART and RF revealed better results in terms of imputation-prediction combination. In the prediction analysis, CART with SVM achieved the highest $R^2$ score among the evaluated pairs of imputation and prediction methods on the test data. RF represented the second-best predictive model, performing properly on the test data imputed with three out of the five imputation methods.

**Funding (optional)**

This work was supported by the Brainteaser project, funded by European Union under Horizon 2020 program (Grant Agreement number: 101017598).